\relax
\documentclass[letterpaper]{article} % DO NOT CHANGE THIS
\usepackage{aaai21}  % DO NOT CHANGE THIS
\usepackage{times}  % DO NOT CHANGE THIS
\usepackage{helvet} % DO NOT CHANGE THIS
\usepackage{courier}  % DO NOT CHANGE THIS
\usepackage[hyphens]{url}  % DO NOT CHANGE THIS
\usepackage{graphicx} % DO NOT CHANGE THIS
\usepackage{natbib}  % DO NOT CHANGE THIS AND DO NOT ADD ANY OPTIONS TO IT
\usepackage{caption} % DO NOT CHANGE THIS AND DO NOT ADD ANY OPTIONS TO IT
\usepackage{amsmath}
\usepackage{amssymb}
\usepackage{array}
\usepackage{adjustbox}
\usepackage{booktabs}

\usepackage{algorithm}
\usepackage{algorithmic}
\usepackage{multirow}
\newcommand{\tabincell}[2]{\begin{tabular}
		{@{}#1@{}}#2
\end{tabular}}

\usepackage[switch]{lineno}  %

\usepackage{float} %-- This package is specifically forbidden
\usepackage{flushend}  %-- This package is specifically forbidden

\title{ProMask: Probability Mask for Skeleton Detection}
\author{
    Xiuxiu Bai, Lele Ye, Zhe Liu \\
}
\affiliations{
    School of Software Engineering, Xi'an Jiaotong University\\
    Xi'an 710049, China\\
    E-mail: xiubai@xjtu.edu.cn; \{yeler082, alfredliu\}@stu.xjtu.edu.cn
}

\begin{document}
\maketitle
% \linenumbers  %

\begin{abstract}
Detecting object skeletons in natural images presents challenging, due to varied object scales, the complexity of backgrounds and various noises. 
The skeleton is a highly compressing shape representation, which can bring some essential advantages but cause the difficulties of detection. This skeleton line occupies a rare proportion of an image and is overly sensitive to spatial position. 
Inspired by these issues, we propose the ProMask, which is a novel skeleton detection model. The ProMask includes the probability mask and vector router. The skeleton probability mask representation explicitly encodes skeletons with segmentation signals, which can provide more supervised information to learn and pay more attention to ground-truth skeleton pixels. Moreover, the vector router module possesses two sets of orthogonal basis vectors in a two-dimensional space, which can dynamically adjust the predicted skeleton position.
We evaluate our method on the well-known skeleton datasets, realizing the better performance than state-of-the-art approaches. Especially, ProMask significantly outperforms the competitive DeepFlux by 6.2\% on the challenging SYM-PASCAL dataset.
We consider that our proposed skeleton probability mask could serve as a solid baseline for future skeleton detection,  since it is very effective and it requires about 10 lines of code.
\end{abstract}

\section{Introduction}
The object skeleton is a low-dimension shape structure representation \cite{Blum1967A}. It can also reflect the connection of various object components \cite{dickinson2009object, Marr1978Representation}. For the deformable object, the skeleton provides a concise and effective representation. Therefore, it can be applied in a variety of fields, such as object recognition and retrieval \cite{zhu1996forms, felzenszwalb2005pictorial, bai2009active}, pose estimation \cite{girshick2011efficient, shotton2011real}, medical diagnosis\cite{naf1996characterization} and shape matching \cite{siddiqi1999shock}. Since tasks such as text detection \cite{zhang2015symmetry} and lane line detection \cite{sironi2014multiscale} in natural scenes are more dependent on the structured information of the object, many visual tasks \cite{zhang2015symmetry,sironi2014multiscale} employ skeleton detection as a pretext task.

%-------------------------------------------------------------------------
\begin{figure}[H]
	\begin{center}
		\includegraphics[width = 1\linewidth]{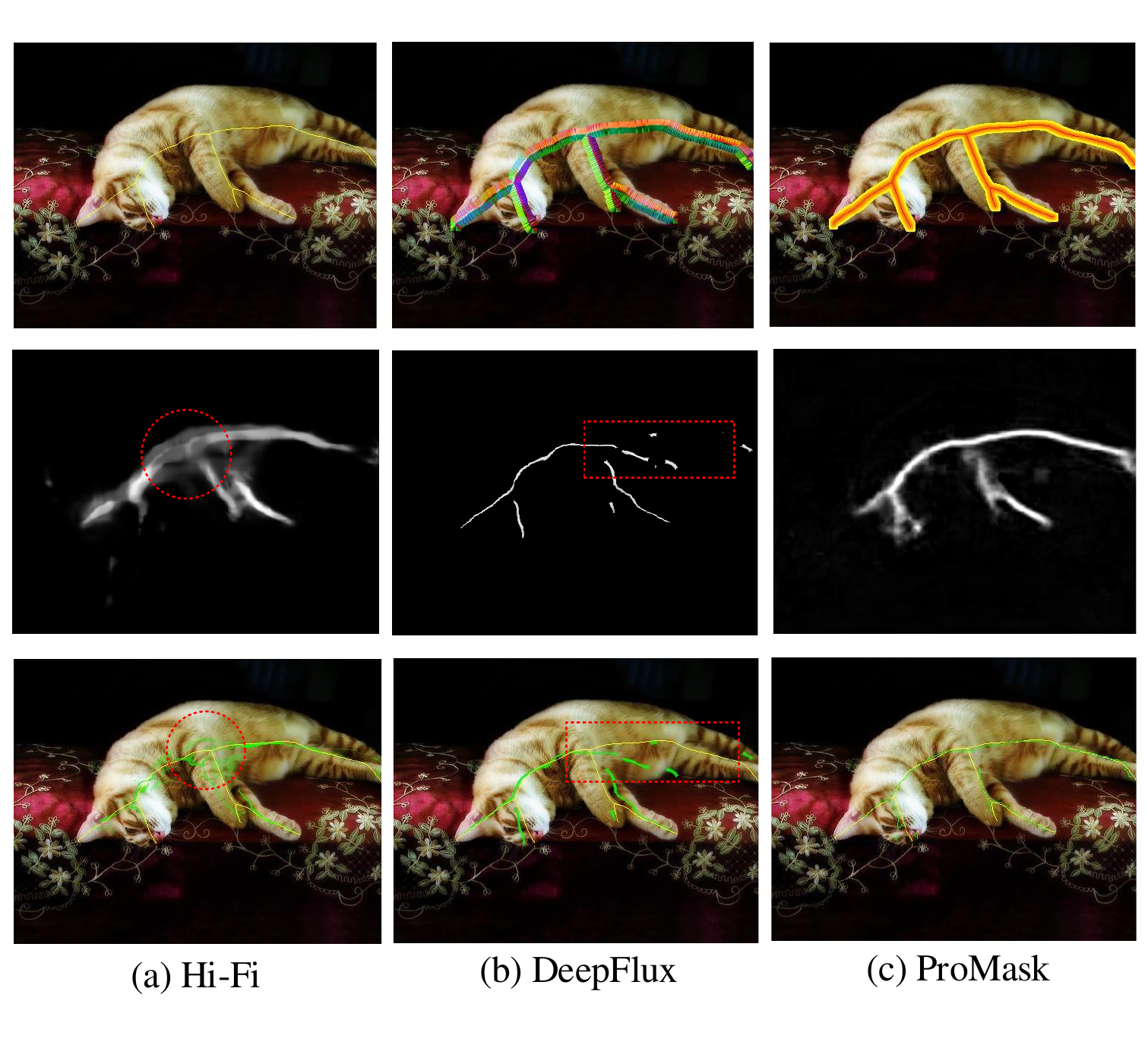}		
	\end{center}
	\caption{Skeleton supervision context comparison. The first row is the supervised labels. The second row is the predicted skeleton by the corresponding methods. (a) Hi-Fi \cite{zhao2018hi}. The original skeleton labels with scales. The predicted skeleton may be disconnected at joints where parts of different scales are connected.  (b) DeepFlux \cite{wang2019deepflux}. The flux vectors are the proxy skeleton labels. Such disconnection and deviation also occurs. (c) The proposed ProMask adopts the skeleton probability mask as the proxy skeleton labels. It can alleviate the above problems. Yellow: ground-truth skeleton; Green: predicted skeleton. }
	\label{fig1}
\end{figure}
%-------------------------------------------------------------------------

Earlier skeleton detection methods \cite{lindeberg1998edge, liu1998segmenting, jang2001pseudo,  yu2004segmentation, nedzved2006gray, zhang2007accurate, lindeberg2013scale} obtain the object skeleton directly through geometric constraints between skeleton pixels and edge information, and then calculate the gradient intensity map. Based on the classic supervised methods \cite{tsogkas2012learning,  shen2016multiple, jerripothula2017object}, the object skeleton detection is first transformed into pixel-level classification or regression problems. Because the effective information such as the shape and texture of the objects in the image are not fully utilized, these methods hardly detect object skeletons in the complex scenes.
%-------------------------------------------------------------------------

Recently convolutional neural networks (CNN) put forward the performance of skeleton detection. Figure~\ref{fig1} shows three supervised skeleton labels. Many classical CNN-based skeleton detection models \cite{zhao2018hi, Shen2016Object, shen2017deepskeleton, Ke2017SRN, liu2017fusing,  liu2018linear} build on the network architecture of nested edge detection (HED) \cite{xie2015holistically}, and directly adopt the ground-truth skeletons as the supervised labels (Figure 1a). These methods focus on exploiting multi-scale and multi-layer information to improve the performance of skeleton detection. DeepFlux \cite{wang2019deepflux} introduces a proxy skeleton flux label, which encodes skeleton pixels with implied object edge pixels (Figure 1b). 

One challenge of skeleton detection is that the skeleton points make up a small part of the image, which raises the difficulty of supervision. In this paper, we introduce a skeleton probability mask representation (Figure 1c), explicitly encoding skeletons with segmentation signals. The probability of closer to true skeleton points is larger. This novel skeleton probability mask can provide sufficient information for detecting skeleton feature, and remain skeleton points as the leading role. Further, we present a vector router module possessing two sets of orthogonal basis vectors in a two-dimensional space, which can dynamically fine-grained adjust the predicted skeleton pixels. The experiments demonstrate that skeleton detection task can benefit from this vector router module.
%-------------------------------------------------------------------------
\begin{figure*}[t]
	\begin{center}
		\includegraphics[width = 0.7\linewidth]{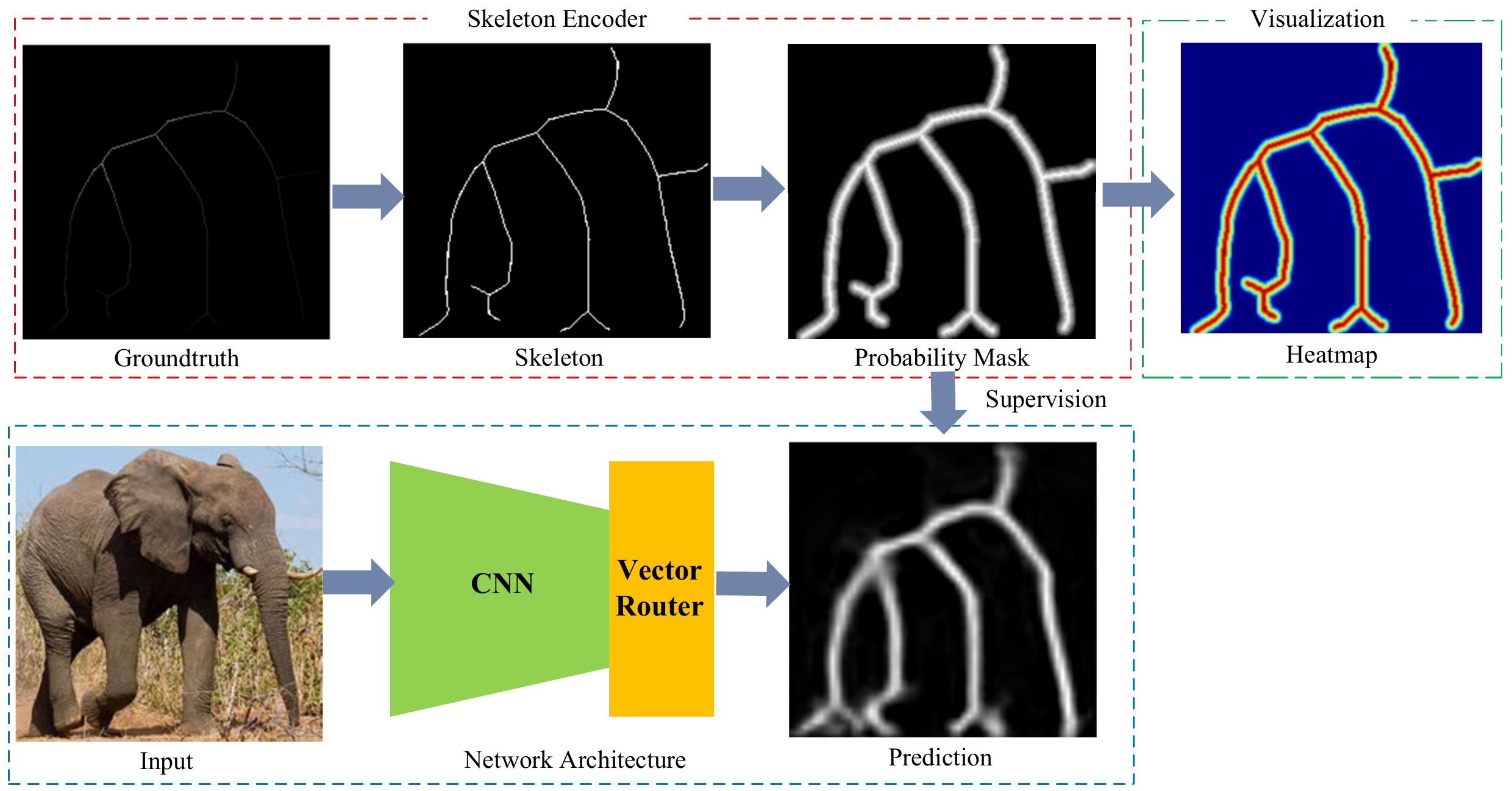}	
	\end{center}
	\caption{The ProMask pipeline. Given a ground-truth label, the model calculates the skeleton probability mask as the supervision label. The heatmap clearly visualizes the probability mask. The vector router module can fine-grained adjust the predicted pixels. }
	\label{fig2}
\end{figure*}

%-------------------------------------------------------------------------
The main contributions of our paper are as follows:

\begin{itemize}

\item[$\bullet$] We introduce the probability mask for skeleton detection, which transforms the traditional skeleton line detection into a soft mask plane segmentation problem. 
This pure skeleton probability mask can achieve a high detection performance.
Due to its simplicity, which can be implemented in about 10 lines of code, we consider our method as a strong baseline for skeleton detection.

\item[$\bullet$] We propose the vector router module, which can fine-grained adjust the predicted pixels to the ground-truth points. The vector router module can promote skeleton detection tasks.

\item[$\bullet$] Our ProMask outperforms the state-of-the-art approaches on most skeleton detection datasets.

\end{itemize}

%-------------------------------------------------------------------------

\section{Related Work}
Since Blum \cite{ Blum1967A} proposed the concept of skeleton feature, a series of skeleton detection methods have been exploited.

\textbf{Unsupervised methods} In the earlier methods \cite{lindeberg1998edge,  liu1998segmenting, jang2001pseudo, yu2004segmentation, nedzved2006gray, zhang2007accurate, lindeberg2013scale}, gradient intensity maps are calculated to directly obtain the object skeleton. Mignotte \cite{mignotte2016symmetry} presents a straight line segments matching to get the skeleton position. Tsogkas et al. \cite{tsogkas2017amat} adopt a weighted geometric set coverage solution to detect object skeletons. Bai et al. \cite{bai2020skeleton} propose a zero-sum self-symmetric Skeleton Filter to obtain the skeleton under highly noisy situations.

\textbf{Classical supervised methods} Tsogkas et al. \cite{ tsogkas2012learning} propose a multiple instance learning (MIL) to train multi-scale and multi-orientation feature vector sets for skeleton detection. Shen et al. \cite{shen2016multiple} expand the scheme of MIL to learn a variety of skeleton patterns. Jerripothula et al. \cite{jerripothula2017object} explore a joint approach that co-processes the skeleton and segmentation problems, exploiting the prior knowledge of object shapes. Sironip et al. \cite{sironi2014multiscale} reconstruct the center line according to the regression problem, and use the local maximum of the regression variable as skeleton positions.

\textbf{Deep learning-based methods} The widespread adoption of deep learning speeds the development of skeleton detection task \cite{zhao2018hi, Shen2016Object, shen2017deepskeleton, Ke2017SRN, liu2017fusing,  liu2018linear}. Many prior skeleton detection methods stem from HED \cite{xie2015holistically} network architecture, which cascaded fuse the last feature map of each block. FSDS \cite{Shen2016Object} further exploits the scale-related skeleton labels to supervise learning, which can precisely capture skeletons at various scales. LMSDS \cite{shen2017deepskeleton} extend to simultaneously learn multi-tasks to promote each other. Ke et al. \cite{Ke2017SRN} introduce a residual module to optimize the skeleton loss between side output and ground truth. Liu et al. \cite{liu2018linear} present a linear span model which exploits linear span module to enhance the independence of feature layers. Hi-Fi \cite{zhao2018hi} leverages hierarchical multi-scale features with bilateral cues, high-level semantics and low-level details. 

DeepFlux \cite{wang2019deepflux} outperforms all the prior methods by introducing the context flux as a substitute learning label, which encodes the skeleton pixels with edge pixels.
Liu et al. \cite{liu2019orthogonal} explore an orthogonal decomposition unit converting a feature map to orthogonal bases, which can reconstruct the skeleton mask.
Xu et al. \cite{xu2019geometry} introduce a new objective function to detect skeleton, which utilizes the Hausdorff distance to achieve the geometry-aware ability.

Bai et al. \cite{bai2020on} first study the robustness of skeleton detection against adversarial attacks.

Considering that skeleton line occupies a rare proportion of an image and skeleton detection is a highly spatial position sensitive task, these attributes raise the difficulty of skeleton supervision. 
To overcome this issue, we introduce a novel skeleton detection model.

%-------------------------------------------------------------------------
\section{Method}
The skeleton feature is a highly compressing shape representation, which can bring some essential advantages but cause the difficulties of detection. 
We propose the \textbf{ProMask}, which is a novel skeleton detection model. Figure~\ref{fig2} depicts the overview of the ProMask including the probability mask and vector router. The skeleton probability mask is used as the proxy supervised labels, which convert the skeleton line to the soft mask plane to strengthen the learned signals. And vector router can be served as a plugin to fine-tune the precise prediction position.

\subsection{Probability Mask}

The probability mask is a weighted mask for skeleton labels. Figure~\ref{fig2} depicts that the visualization of skeleton probability mask. The principle behind probability mask is that the weight of position closer to true skeleton is larger. 

We formally define an image \emph{X}.
1)	\emph{B} denotes the background pixels. $B_{i} = 0$;
2)	\emph{S} denotes the skeleton pixels of an image. $S_{i} \in (0, 255)$, the larger values represent the skeleton pixels with larger scales, as shown in Figure~\ref{fig2}. We calculate the skeleton probability mask by Algorithm 1.

%Algorithm~\ref{alg:1} shows how to calculate the skeleton probability mask labels. The inputs are ground-truth skeleton labels \emph{S} and mask radius \emph{r}. 
%First, it binarizes $S$.
%Second, it dilates the skeleton labels with \emph{r} to get the dilated mask $S_{dilate}$. $S_{dilate}$ denotes that the weights of all dilated skeleton pixels are 1. 
%Third, it makes the $S_{prob}$ to obey the Gaussian distribution by using the $GaussianBlur()$ function, then normalize. 
%Finally, we obtain the skeleton probability mask by adding the normalization of original skeletons with the weighted mask. 
%The purpose of adding the normalization of original skeletons is to pay more attention on the large-scale parts of objects. 
%$dilate()$ and $GaussianBlur()$ functions are from OpenCV libarary.

Algorithm~\ref{alg:1} shows how to calculate the skeleton probability mask labels. The inputs are ground-truth skeleton labels \emph{S} and mask radius \emph{r}. 
First, it binarizes $S$.
Second, it dilates the skeleton labels with \emph{r} to get the dilated mask $S_{dilate}$. $S_{dilate}$ denotes that the weights of all dilated skeleton pixels are 1. 
Third, it makes the $S_{prob}$ to obey the Gaussian distribution by using the $GaussianBlur()$ function, then normalize. Fourth, it obtains skeleton joints by using a classical morphological function.
Finally, we obtain the skeleton probability mask by adding the normalization of original skeletons and joints with the weighted mask. 
The aim of adding the normalization of original skeletons is to pay more attention on the large-scale parts of objects. 
The aim of adding joints is to more focus on object joints, which are easily disconnected.
$dilate()$ and $GaussianBlur()$ are from OpenCV libarary.

%===================================

Theoretically, this skeleton probability mask make the network to have more supervised information to learn, and simultaneously pay more attention to ground-truth skeleton positions. We call this mode the medial axis attention mechanism, which enables the skeleton detection model to become more robust. Experimentally, our skeleton probability mask leads to big improvements in such complex cases. 

%-------------------------------------------------------------------------
\begin{algorithm}
	\caption{Encoding skeleton labels to Probability Mask}
	\begin{algorithmic}[1]
		\REQUIRE  
		Skeleton of image $S$, Mask radius $r$\\
		\ENSURE Skeleton Probability Mask $S_{proMask}$ 
		\STATE $//$ binarize
		\STATE $S_{bin} \leftarrow binarize(S) $
		
		\STATE $//$ compute the skeleton probability mask
		\STATE $kernel \leftarrow ones(r, r)$
		\STATE $S_{dilate} \leftarrow dilate(S_{bin}, kernel)$
		\STATE $S_{prob} \leftarrow GaussianBlur(S_{dilate}, (r, r) )$
		
		\STATE $//$  compute skeleton joints
		\STATE $joint \leftarrow bwmorph(S_{bin})$
		\STATE $joint \leftarrow dilate(joint) * S_{bin}$
		
		\STATE $//$  normalize
		\STATE $S_{prob} \leftarrow normalize(S_{prob}) $
		\STATE $S \leftarrow normalize(S)$
		\STATE $S_{proMask} \leftarrow  S_{prob} + S + joint$
		\STATE $S_{proMask} \leftarrow normalize(S_{proMask}) $
		\RETURN $S_{proMask}$
	\end{algorithmic}
	\label{alg:1}
\end{algorithm}
%-------------------------------------------------------------------------

%%-------------------------------------------------------------------------
%\begin{algorithm}
%	\caption{Encoding skeleton labels to Probability Mask}
%	\begin{algorithmic}[1]
%		\REQUIRE  
%		Skeleton of image $S$, Mask radius $r$\\
%		\ENSURE Skeleton Probability Mask $S_{proMask}$ 
%		\STATE $//$ binarize
%		\STATE $S_{bin} \leftarrow binarize(S) $
%		
%		\STATE $//$ compute the skeleton probability mask
%		\STATE $kernel \leftarrow ones(r, r)$
%		\STATE $S_{dilate} \leftarrow dilate(S_{bin}, kernel)$
%		\STATE $S_{prob} \leftarrow GaussianBlur(S_{dilate}, (r, r) )$
%		
%		\STATE $//$  normalize
%		\STATE $norm\_S_{prob} \leftarrow normalize(S_{prob}) $
%		\STATE $norm\_S \leftarrow normalize(S)$
%		\STATE $S_{proMask} \leftarrow  norm\_S_{prob} + norm\_S $
%		\RETURN $S_{proMask}$
%	\end{algorithmic}
%	\label{alg:1}
%\end{algorithm}
%%-------------------------------------------------------------------------

\subsection{Vector Router}
Skeleton detection is highly sensitive to spatial position. The predicted skeleton points may be deviated or even disconnected caused by joints or kinds of perturbations. To overcome this issue, we propose a vector router module, which can fine-tune the predicted position. The basic block of vector router is a pair of orthogonal vector kernels $(1 \times n, n \times 1)$. It can involve $n \times n$ neighbor signals to assist adjusting the ill-suited outputs.  

\begin{figure}[t]
	\begin{center}
		\includegraphics[width = 0.7\linewidth]{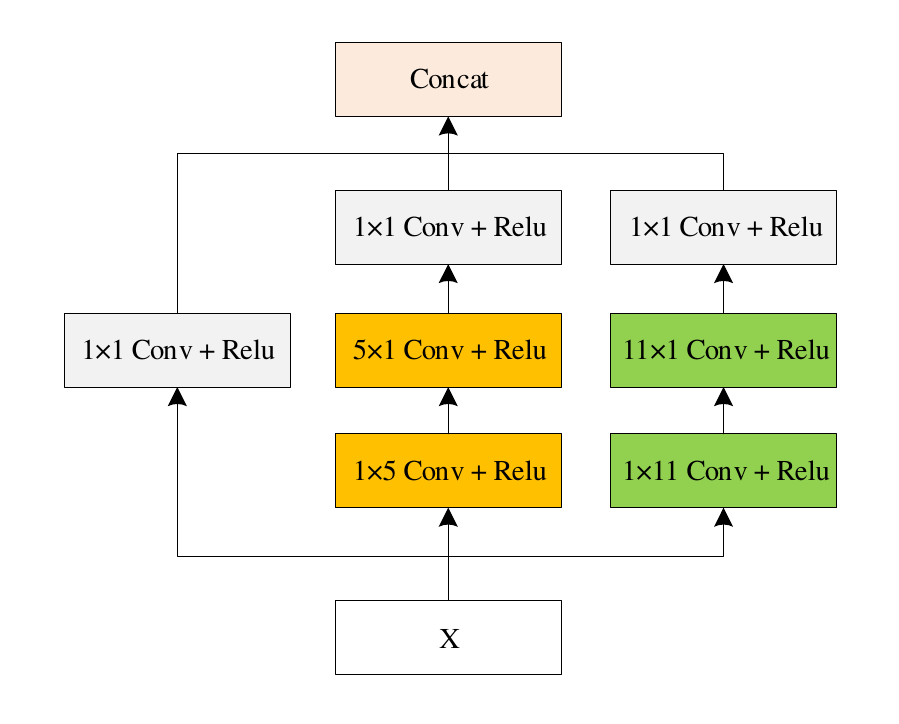}	
	\end{center}
	\caption{One version of the vector router module. It consists of two pairs of orthogonal vector convolutional kernels $(1 \times 5, 5 \times 1)$ and $(1 \times 11, 11 \times 1)$ and a residual connection. The fusion operations include concatenation and $1 \times 1$ convolution. It can use the Inception-like module to directly achieve this fine-tuned function.}
	\label{fig3}
\end{figure}

Figure~\ref{fig3} shows one version of vector router. We adopt two pairs of orthogonal vector convolutional kernels $(1 \times 5, 5 \times 1)$ and $(1 \times 11, 11 \times 1)$. Moreover, we add a branch of $1 \times 1$ convolution as a residual connection. The number of channels of these three branches are set as (256, 256, 256). The output of vector router uses $1 \times 1$ convolution to fuse the feature maps.

\begin{figure}[t]
	\begin{center}
		\includegraphics[width = 0.95 \linewidth]{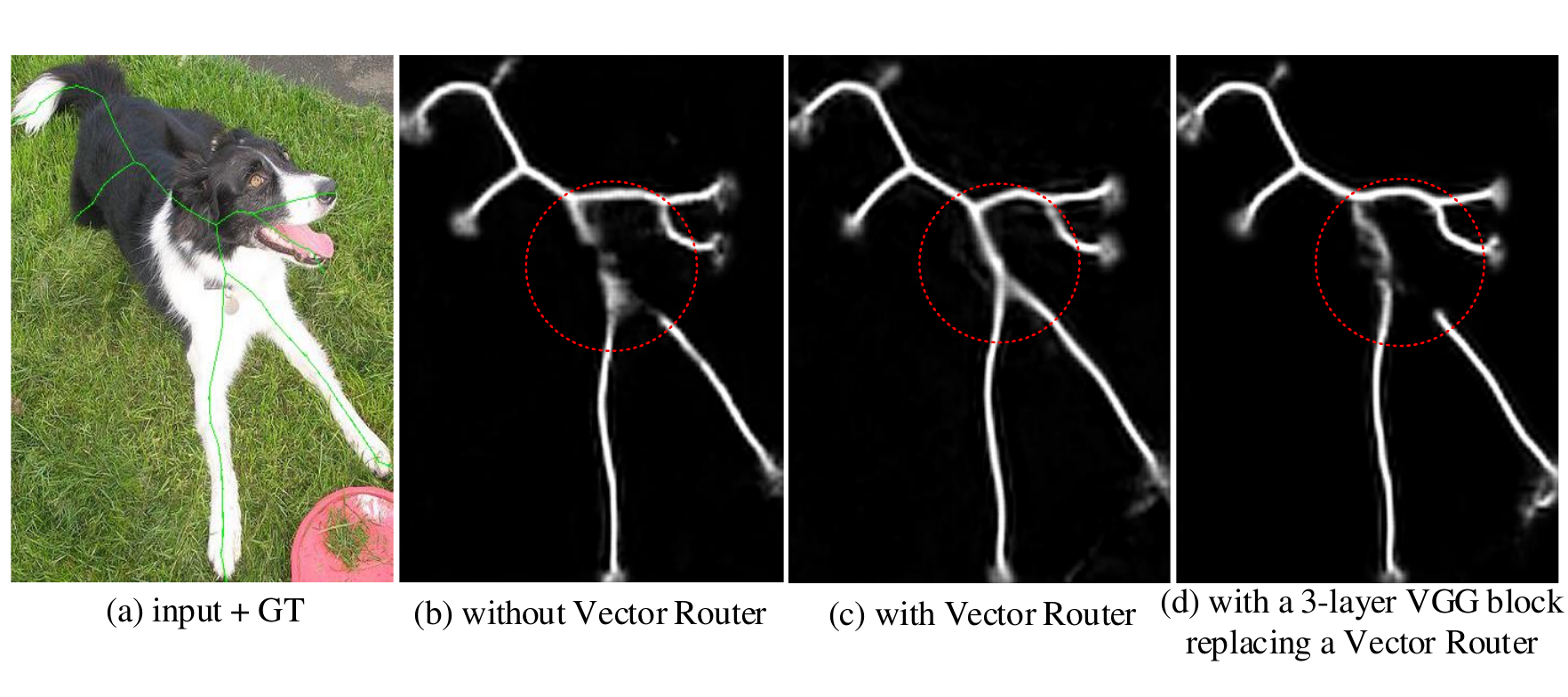}		
	\end{center}
	\caption{The effect of vector router. It can compensate for the lost feature by involving the neighbor signals.}
	\label{fig4}
\end{figure}

Figure~\ref{fig4} visually compares the effect of vector router. The red circle highlights the disconnected skeleton lines without using vector router.
It is easy to occur at joints where the appearance changes greatly.
While the vector router can strengthen the weak skeleton signals by involving the neighbor skeleton pixels. 
If we use a 3-layer VGG block to replace a vector router at the same location, it cannot work well, as shown in Figure~\ref{fig4}(d).
The possible reason is that a $n \times n$ filter involves too much background pixels dominating the prediction.
While a single separate $1 \times n$ or $n \times 1$ filter gives neighbor skeleton pixels a chance to make up for the disconnected joints (in Figure~\ref{fig4}(c)).
Therefore, it can improve the performance by employing a vector router module.

%-------------------------------------------------------------------------
%-------------------------------------------------------------------------
%-------------------------------------------------------------------------
%-------------------------------------------------------------------------

\subsection{Architecture of ProMask}
%-------------------------------------------------------------------------
\begin{figure}[t]
	\begin{center}
		\includegraphics[width = 0.95\linewidth]{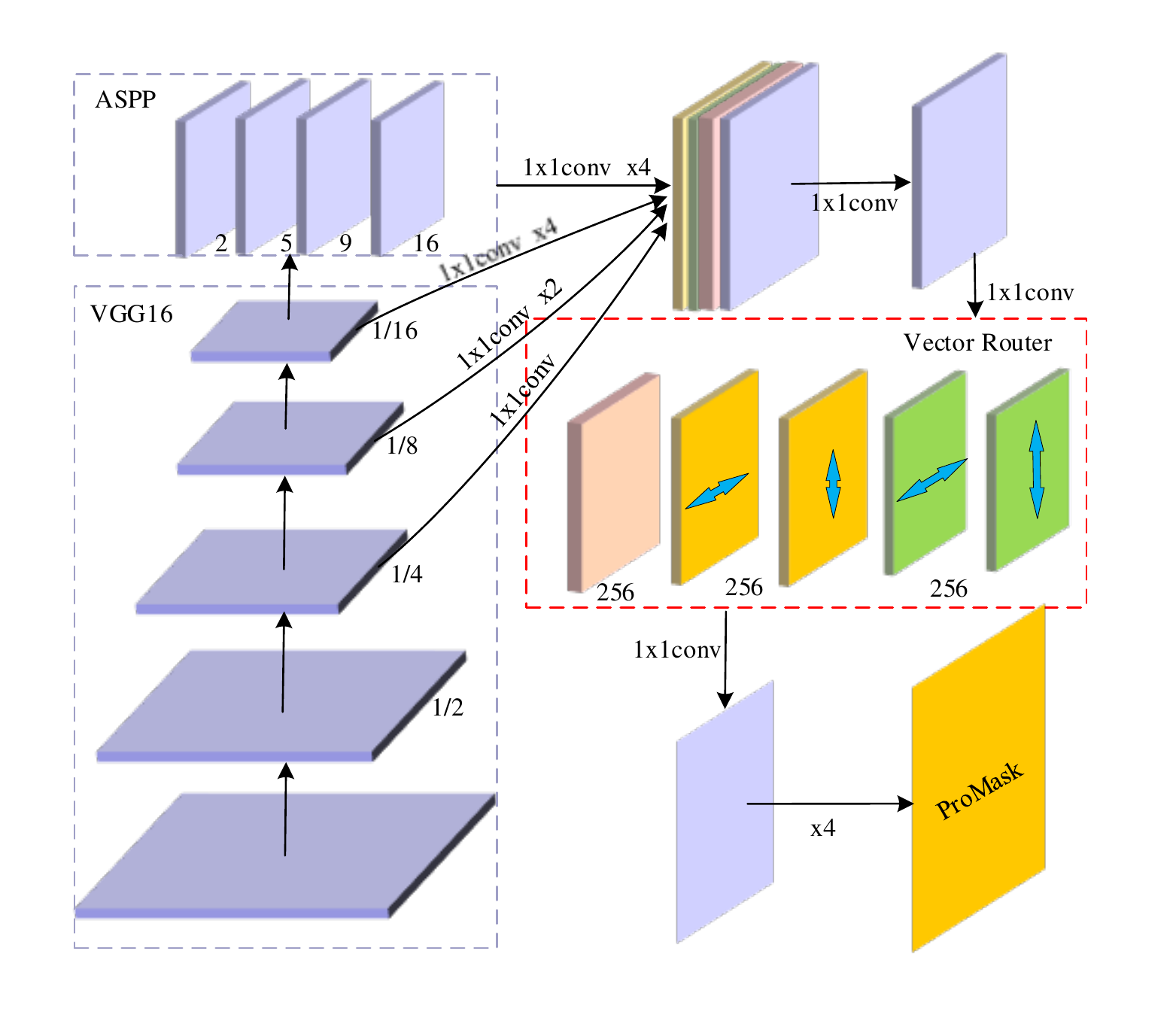}
	\end{center}
	\caption{The ProMask architecture. The used backbone network is the VGG16 \cite{simonyan2014very} with ASPP \cite{chen2017deeplab}. We design a vector router module to predict a 1-channel skeleton probability mask.}
	\label{fig5}
\end{figure}
%-------------------------------------------------------------------------

Figure~\ref{fig5} shows the network architecture of our method. It includes five modules: 1) a backbone feature extraction network VGG16 \cite{simonyan2014very}; 2) an atrous spatial pyramid pooling (ASPP) module \cite{chen2017deeplab}; 3) a multi-level feature fusing module; 4) a vector router module; 5) the skeleton probability mask for learned labels.

For a fair comparison with existing methods, we choose VGG16 \cite{simonyan2014very} as the backbone architecture. In the ASPP module, we adopt the dilation rate (2, 5, 9, 16) to alleviate “gridding effect” caused by the common dilation rate (2, 4, 8, 16) \cite{wang2018understanding}. 

Since the skeleton feature is sensitive to object scales, we explore a multi-level feature fusing module, which combines the feature maps of conv3\_3, conv4\_3, conv5\_3 and ASPP output. It uses bilinear interpolation to rescale the same size before concatenating. Next, 1 $\times$ 1 convolution is applied to conduct the feature interaction. The up-sampling operation may cause the aligned problem. Therefore, we design a vector router module to adjust the predicting pixels. The last output of the model is a 1-channel skeleton probability mask.

%-------------------------------------------------------------------------
\begin{table*}[t]
	\begin{center}
		\renewcommand{\arraystretch}{1}
		\begin{adjustbox}{width=\linewidth}
        \scriptsize
		\begin{tabular}{lccccc}
			\toprule
			Method & SK-LARGE & SK506 &WH-SYMMAX & SYM-PASCAL & SYMMAX300\\ \hline 
			MIL \cite{tsogkas2012learning} & 0.353 & 0.392 & 0.365  & 0.174 & 0.362\\ 		
			HED \cite{xie2015holistically} & 0.497 & 0.541 & 0.732  & 0.369 & 0.427\\ 
			RCF \cite{liu2017richer} & 0.626 & 0.613 & 0.751  & 0.392 & -  \\  
			FSDS \cite{Shen2016Object} & 0.633 & 0.623 & 0.769  & 0.418 & 0.467 \\ 
			LMSDS \cite{shen2017deepskeleton} & 0.649 & 0.621 & 0.779  & - & -\\ 
			SRN \cite{Ke2017SRN} & 0.678 & 0.632 & 0.780  & 0.443 & 0.446 \\ 
			OD-SRN \cite{liu2019orthogonal} & 0.676 & 0.624 & 0.804  & 0.444 & 0.489 \\ 
			LSN \cite{liu2018linear} & 0.668 & 0.633 & 0.797  & 0.425 & 0.480 \\ 
			Hi-Fi \cite{zhao2018hi} & 0.724 & 0.681 & 0.805  & 0.454 & -\\
			DeepFlux \cite{wang2019deepflux} & 0.732 & 0.695 & 0.840  & 0.502 & 0.491 \\
			GeoSkeletonNet* \cite{xu2019geometry} & \textbf{0.757} & \textbf{0.727} & 0.849  & 0.520 & 0.501 \\ \hline
			ProMask (PM) (Ours) & 0.739 & 0.701 & 0.849  & 0.555 & \textbf{0.540} \\ 
			ProMask (PM + VR) (Ours)  & 0.748 & 0.711 & \textbf{0.858}  & \textbf{0.564} & 0.525 \\ 
			\bottomrule   
		\end{tabular}
	\end{adjustbox}
	\end{center}
	\caption{F-measure comparison on five skeleton datasets. The larger is better. PM: probability mask. VR: vector router. The competing results are from the respective papers. *denotes using average gradients and resolution normalization. For GeoSkeletonNet, the 2.9\% performance improvement comes from training strategies including average gradients and resolution normalization on the SK-LARGE dataset. While other skeleton detection methods do not use these training strategies. 
	}
	\label{table:1}
\end{table*}
%-------------------------------------------------------------------------

\subsection{Loss Function}
We use the weighted $L_{2}$ loss to train the network. In an image $X$, skeleton pixels $S$ are much less than background pixels $B$, which cause the unbalance problem. Although we adopt the skeleton probability mask to supervise, this unbalance problem also exists. Therefore, we utilize the weighted $L_{2}$ loss as 
\begin{equation}
L=\sum_{i=1}^{N}w(i)*\left \| X(i) ,\right. P(i)\|_{2}
\end{equation}
where $i \in (1,...,N)$ denotes each pixels. $P(i)$ denotes the predicted values. The balance weight $w(i)$ is as follows:
\begin{equation}
	w(i)= \left\{\begin{matrix}
	\frac{S^{(n)}}{X^{(n)}} &,X^{(i)} = 0 \\ 
	 \frac{B^{(n)}}{X^{(n)}} & ,X^{(i)} > 0
	 \end{matrix}\right.
\end{equation}
where $X^{(n)}$, $S^{(n)}$ and $B^{(n)}$ represent the number of image, skeleon probability mask and background pixels.
%-------------------------------------------------------------------------

\section{Experiments}
We evaluate our method on five challenging skeleton datasets: SK-LARGE \cite{shen2017deepskeleton}, SK506 \cite{Shen2016Object}, WH-SYMMAX \cite{shen2016multiple}, SYM-PASCAL \cite{Ke2017SRN}, and SYMMAX300 \cite{tsogkas2012learning}.

\subsection{Datasets}
\textbf{SK-LARGE} \cite{shen2017deepskeleton} is selected from the MS COCO dataset \cite{chen2015microsoft}, which contains 746 training images and 745 testing images. 

\textbf{SK506} \cite{Shen2016Object} is an old version of SK-LARGE, which includes 300 training images and 206 testing images.

\textbf{WH-SYMMAX} \cite{shen2016multiple} is stemmed from the Weizmann Horse dataset \cite{borenstein2002class}, which contains 227 training images and 100 testing images.

\textbf{SYM-PASCAL} \cite{Ke2017SRN} is built on the PASCAL VOC segmentation dataset \cite{everingham2010pascal}, which contains 647 training images and 788 testing images. 

\textbf{SYMMAX300} \cite{tsogkas2012learning} is from the BSDS300 dataset \cite{martin2001database}. It consists of 200 training images and 100 testing images.

Among these datasets, SK-LARGE, SK506 and WH-SYMMAX focus on single object of cropped images for skeleton detection, while SYM-PASCAL and SYMMAX300 contain multiple objects of an image, which can raise the barrier of skeleton detection. 

\subsection{Evaluation Metric}
We adopt the F-measure metric and precision-recall (PR) curves to quantitatively verify the preformation. F-measure is calculated by the formula $F = 2PR / (P + R)$. The skeleton detection methods use non-maximal suppression (NMS) \cite{dollar2014fast} to get the skeleton map and then use threshold and the thinning algorithm \cite{nemeth2011thinning} to obtain the final thinned skeletons. 

%-------------------------------------------------------------------------

\subsection{Implementation Details}

Our network architecture is based on the Pytorch framework. The evaluating platform is an Intel (R) Xeon (R) CPU (2.60GHz), 64GB RAM, on a single GTX 2080Ti GPU.

For a fair comparison, we use the same data augmentation as DeepFlux \cite{wang2019deepflux}. The specific strategies are: 1) rescale the input image to three input scales (0.8, 1.0, 1.2); 2) flip the input image from left to right and up and down; 3) rotate input image to four directions $(0^\circ, 90^\circ, 180^\circ, 270^\circ)$.

To obtain the probability mask, we set a mask radius parameter $r$. We use $r = 7$ for the SYM-PASCAL dataset and $r = 5$ for other datasets. In the training, we adopt the VGG16 \cite{simonyan2014very} network as the backbone, and then fine-tune on the pretrained ImageNet \cite{deng2009imagenet} model to finally obtain a model for skeleton detection. We use ADAM \cite{ kingma2014adam} as an optimizer. To mitigate over-fitting, we set a weight decay as 0.015. The model first iterates 80k on a 1e-4 learning rate, and then reduces the learning rate to 1e-5 for the next 20k iterations. Our code will be open source.

%-------------------------------------------------------------------------
\begin{figure}[t]
	\begin{center}
		\includegraphics[width = 1\linewidth]{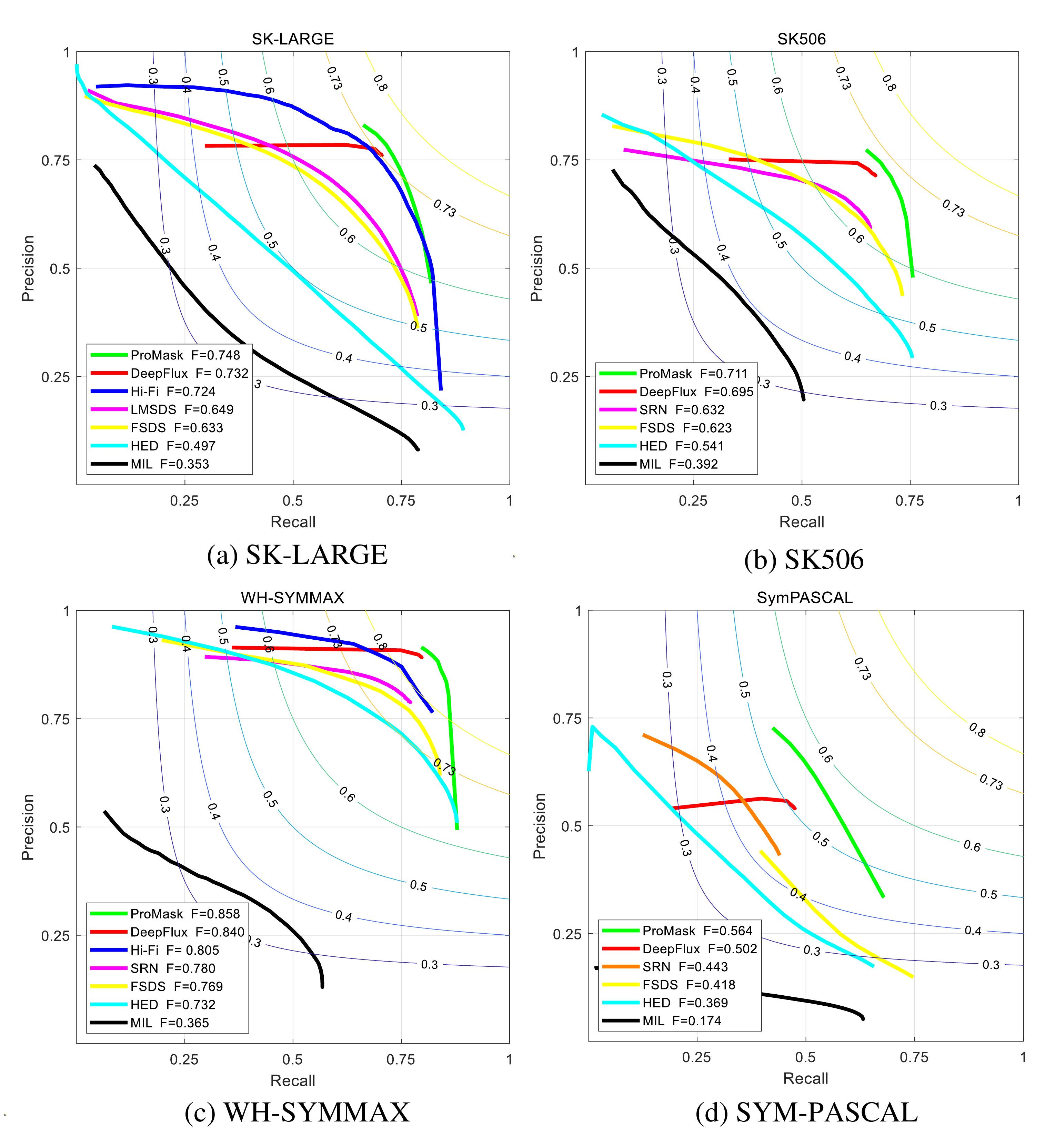}		
	\end{center}
	\caption{PR curves of the popular methods for skeleton detection. ProMask offers high recall. }
	\label{fig6}
\end{figure}

%-------------------------------------------------------------------------
%-------------------------------------------------------------------------
\begin{figure*}[t]
	\begin{center}
		\includegraphics[width = 0.91 \linewidth]{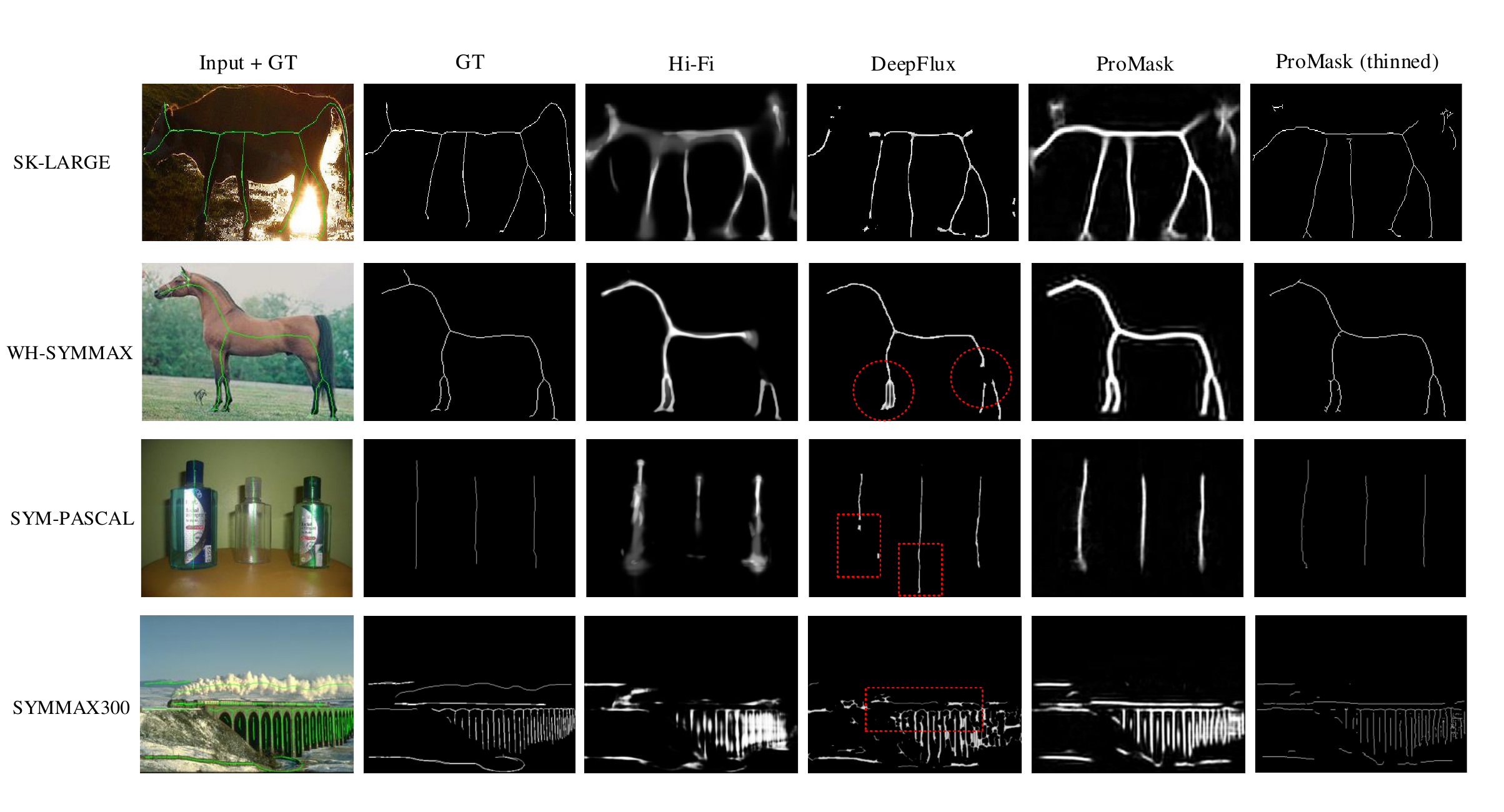}
	\end{center}
	\caption{Qualitative comparison with the current competitive methods Hi-Fi and DeepFlux on the four skeleton datasets. Our ProMask performs better than the previous methods, particularly in the joint parts and complex situations. }
	\label{fig7}
\end{figure*}

%-------------------------------------------------------------------------

%-------------------------------------------------------------------------
\begin{table}[t]
	\begin{center}
		\renewcommand{\arraystretch}{1.1}
		\begin{adjustbox}{width=\linewidth}
        \scriptsize
		\begin{tabular}{lcc}
			\toprule
			Method & F-measure & Runtime (sec) \\ \hline  			
			FSDS \cite{ Shen2016Object} & 0.418 &  0.019 \\ 
			Hi-Fi \cite{zhao2018hi} & 0.454 & 0.054 \\ 
			DeepFlux \cite{wang2019deepflux} & 0.502 &  0.015 \\ \hline
			ProMask (PM) (Ours) & \textbf{0.555} &  \textbf{0.011} \\ 
			ProMask (PM + VR) (Ours) & 0.564 &  0.014 \\ 
			\bottomrule 
		\end{tabular}
	\end{adjustbox}
	\end{center}
	\caption{Inference time and performance on SYM-PASCAL.}
	\label{table:2}
\end{table}

%-------------------------------------------------------------------------

\subsection{Results}
Table~\ref{table:1} clearly shows the results of all classical skeleton detection methods. Our ProMask achieves the state-of-the-art performance under the same augmentation strategy. Particularly, ProMask outperforms DeepFlux \cite{wang2019deepflux} by 6.2\% and 3.4\% on SYM-PASCAL and SYMMAX300, respectively. 
 ProMask also performs better than DeepFlux by 1.6\%, 1.6\% and 1.8\% on SK-LARGE, SK506 and WH-SYMMAX, respectively. ProMask significantly surpasses Hi-Fi \cite{zhao2018hi} by a large margin 11.0\% on the challenging SYM-PASCAL.

The performance improvement for the first two datasets is more significant than the last three ones.
It is related to the characteristics of datasets. 
The SK-LARGE, SK506 and WH-SYMMAX datasets contain single object of cropped images, which are relatively easier to detect skeleton. The skeleton detection performance is already high. Thus, the performance improvement is limited.
While the SYM-PASCAL and SYMMAX300 datasets are more diverse and challenge due to multiple objects in a single image and complex backgrounds. Therefore, our approach performs much better.

According to the ablation study of GeoSkeletonNet \cite{xu2019geometry}, the 2.9\% performance improvement on the SK-LARGE dataset comes from training strategies, including average gradients and resolution normalization. While other skeleton detection methods do not use these training strategies. 
Various image resolutions increase the difficulties of skeleton detection, but a robust skeleton detection model should have the ability to directly process the input images with different resolutions.

\begin{table}[t]
	\begin{center}
		\renewcommand{\arraystretch}{1.2}
			\begin{adjustbox}{width=\linewidth}
        		\scriptsize
			\begin{tabular}{c|c|c|c|c|c}
			\toprule
			Dataset   & \multicolumn{3}{c|}{\tabincell{c}{Supervision labels: \\ Probability Mask}} & \tabincell{c}{Vector \\ Router} & \tabincell{c}{F-\\measure}  \\  \cline{2-4}
					\multirow{2}*{~} 	& \tabincell{l}{$P=1$\\$r=1$}&  \tabincell{l}{$P=1$\\$r=7$} & \tabincell{l}{$P=S_{proMask}$\\$r=7$} & \multirow{2}*{~} &  \multirow{2}*{~}\\ \hline
			\multirow{5}*{SK-LARGE}   & \checkmark &             &            &            &0.705 \\ 
						 ~        & \checkmark &              &            & \checkmark  &0.711 \\
				             ~        &            &  \checkmark &            &            &0.674 \\
			                 ~        &            &             & \checkmark &            & \textbf{0.735}\\ 
			                 ~        &            &             & \checkmark  & \checkmark & \textbf{0.745}\\ \hline
			
			\multirow{5}*{SYM-PASCAL} & \checkmark &             &            &            & 0.436 \\
							~		 & \checkmark &             &            &\checkmark  & 0.504 \\
			      		     ~        &  		   &  \checkmark &            &            &0.492 \\
			 	             ~        &            &             & \checkmark &            & \textbf{0.555} \\ 
			                 ~        &            &             & \checkmark & \checkmark & \textbf{0.564} \\ 
			\bottomrule 
		\end{tabular}
	\end{adjustbox}
	\end{center}
	\caption{The effect of the skeleton probability mask and vector router on the F-measure. The single skeleton probability mask can greatly improve the performance, which could serve as a strong baseline for future skeleton detection. }
	\label{table:3}
\end{table}

%-------------------------------------------------------------------------
Figure~\ref{fig6} plots the PR curves of all methods. Our ProMask clearly outperforms the comparative approaches. ProMask offers high recall. Figure~\ref{fig7} shows the qualitative results of ProMask, DeepFlux and Hi-Fi. It demonstrates that the outputs of ProMask possess a uniform width like DeepFlux and avoid the skeleton loss of object junctions. For the complex skeleton datasets, ProMask can effectively detect skeletons under noisy backgrounds.

\subsection{Runtime analysis}
ProMask is an end-to-end skeleton detection model, which does not need the post-processing stage. The inference time of  probability mask only requires about 0.011 seconds for a 300 $\times$ 200 image on a single GPU. Our ProMask possesses a significant advantage in terms of performance and efficiency, as shown in Table~\ref{table:2}.

%-------------------------------------------------------------------------

\begin{table}[t]
	\begin{center}
		\renewcommand{\arraystretch}{1.15}
			\begin{adjustbox}{width=\linewidth}
        		\scriptsize
		\begin{tabular}{cccccc}
			\toprule
			Dataset & $r = 3$ & $r = 5$ & $r = 7$ & $r = 9$ & $r = 11$ \\ \hline 		
			SK-LARGE & 0.738 &  \textbf{0.748} & 0.745 & 0.738 & 0.737 \\ 
			SYM-PASCAL & 0.545 &  0.558 & \textbf{0.564} & 0.559 & 0.558\\ 
			\bottomrule 
		\end{tabular}
		\end{adjustbox}
	\end{center}
	\caption{The effect of the mask radius of probability mask on the F-measure.}
	\label{table:4}
\end{table}

\begin{table}[t]
	\begin{center}
		\renewcommand{\arraystretch}{1.15}
			\begin{adjustbox}{width=\linewidth}
        		\scriptsize
		\begin{tabular}{cccc}
			\toprule
			Dataset & $branch = 1$ & $branch = 2$ & $branch = 4$  \\ \hline 		
			SK-LARGE & 0.744  & \textbf{0.745}  & 0.738 \\ 
			SYM-PASCAL & 0.560  & \textbf{0.564} & 0.556\\ 
			\bottomrule 
		\end{tabular}
		\end{adjustbox}
	\end{center}
	\caption{The effect of the connection mode of vector router on the F-measure. $branch$ denotes that the branches  $(1 \times 5),  (5 \times 1)$, $(1 \times 11)$ and $(11 \times 1)$ are connected in series or parallel. $branch = 1$ denotes all branches are in series and $branch = 4$ denotes all branches are in parallel. Here, it uses $r = 7$.}
	\label{table:5}
\end{table} 

%-------------------------------------------------------------------------

\subsection{ Ablation study }
We conduct the ablation study on the representative datasets SK-LARGE and SYM-PASCAL. Table~\ref{table:3} lists three kinds of supervised labels: 

Case1: Probability Mask $(P = 1, r = 1)$ means the traditional standard skeleton labels.

Case2: Probability Mask $(P = 1, r = 7)$ denotes skeletons with mask radius 7 and the weights of all dilated skeleton pixels are 1.

Case3: Probability Mask $(P = S_{proMask}, r = 7)$ represents skeletons with mask radius 7 and the weights of dilated skeleton pixels are obeyed with Gaussian distribution.

The baseline network adopts the skeleton labels in Case 1. It can achieve the performance of 0.711 and 0.504 on SK-LARGE and SYM-PASCAL. Obviously, the probability mask of Case 3 is more accurate than Cases 1 and 2 by 2.4\% and 6.1\% on SK-LARGE. Case 3 outperforms Cases 1 and 2 by 5.1\% and 6.3\% on SYM-PASCAL.
Moreover, we add the vector router on Case 3. It can further improve 1.0\% and 0.9\% on SK-LARGE and SYM-PASCAL. The ablation experiments verify the advantages of our two modules.  

The predicted skeleton may be disconnected at joints where different scale parts are connected. It could be caused by the fusion stage of different scale feature maps.
The skeleton probability mask obeyed with Gaussian distribution can make the skeleton position more prominet and take into account the extra segmentation information.
It can make outputs of ProMask possess a uniform width to avoid the loss of object junctions.
The vector router enhances the weak skeleton joints by involving neighbor skeleton signals.
Therefore, our method can greatly improve skeleton detection results.

The probability mask needs to set a mask radius $r$. Table~\ref{table:4} shows the results of a series of $r$ on SK-LARGE and SYM-PASCAL. $r = 7$ can obtain the best performance on the SYM-PASCAL dataset, and $r = 5$ is best for other datasets. The suitable probability mask radius is beneficial for leveraging more supervised information, while a too large mask radius may disturb other objects or parts in the same image.

Table~\ref{table:5} shows the effect of the connection mode of vector router. $branch = 2$ denotes that the branches  $(1 \times 5, 5 \times 1)$ and $(1 \times 11, 11 \times 1)$ are connected in parallel (as in Figure~\ref{fig3}). This connection mode can effectively involve the neighbor signals to assist adjusting the deviated outputs.

\section{Conclusion}
In this paper, we introduce the skeleton probability mask that explicitly encodes the skeleton pixels with segmentation signals. 
This single module can greately promote the performance of skeleton detection. 
Moreover, we propose a vector router module that dynamically fine-grained adjusts the prediction position by utilizing neighbor skeleton signals. The evaluation on the challenging datasets verifies that ProMask achieves the state-of-the-art performance. In the future, we will explore the robustness of skeleton detection.

\section{ Acknowledgments}
This work was supported by the National Natural Science Foundation of China under Grants 61802297.

\bibliographystyle{aaai21}
\bibliography{egbib}

\end{document}